\documentclass[runningheads]{llncs}
\usepackage{hyperref}
\usepackage{graphicx}
\usepackage{booktabs}
\usepackage{multirow}
\usepackage{subfig}
\usepackage{amsmath}
\usepackage[belowskip=-15pt,aboveskip=0pt]{caption}

\DeclareSymbolFont{toneletters}{T1}{\familydefault}{m}{it}
\SetSymbolFont{toneletters}{bold}{T1}{\familydefault}{bx}{it}

\DeclareMathSymbol\ethm{\mathord}{toneletters}{"F0}
\begin{document}
\title{SwinDocSegmenter: An End-to-End Unified Domain Adaptive Transformer for Document Instance Segmentation}
%
\titlerunning{SwinDocSegmenter}
%
\author{Ayan Banerjee$^*$\inst{1}\orcidID{0000-0002-0269-2202} \and
Sanket Biswas$^*$\inst{1}\orcidID{0000-0001-6648-8270} \and
Josep Lladós\inst{1}\orcidID{0000-0002-4533-4739} \and
Umapada Pal\inst{2}\orcidID{0000-0002-5426-2618}}
\authorrunning{A.Banerjee et al.}
\institute{Computer Vision Center \& Computer Science Department  \\
              Universitat Autònoma de Barcelona, Spain \\
              \email{\{abanerjee, sbiswas, josep\}@cvc.uab.es} 
              \and
              CVPR Unit, Indian Statistical Institute, India \\ 
              \email{umapada@isical.ac.in}}
%
%
%
\maketitle              
\def\thefootnote{*}\footnotetext{These authors contributed equally to this work.}
\begin{abstract}
Instance-level segmentation of documents consists in assigning a class-aware and instance-aware label to each pixel of the image. It is a key step in document parsing for their understanding. In this paper, we present a unified transformer encoder-decoder architecture for en-to-end instance segmentation of complex layouts in document images. The method adapts a contrastive training with a mixed query selection for anchor initialization in the decoder. Later on, it performs a dot product between the obtained query embeddings and the pixel embedding map (coming from the encoder) for semantic reasoning. Extensive experimentation on competitive benchmarks like PubLayNet, PRIMA, Historical Japanese (HJ), and TableBank demonstrate that our model with SwinL backbone achieves better segmentation performance than the existing state-of-the-art approaches with the average precision of \textbf{93.72}, \textbf{54.39}, \textbf{84.65} and \textbf{98.04} respectively under one billion parameters. The code is made publicly available at: \href{https://github.com/ayanban011/SwinDocSegmenter}{github.com/ayanban011/SwinDocSegmenter}


\end{abstract}
%
%
%

\section{Introduction}
\label{s:intro}
Document Intelligence (DI) systems help to provide solutions for automating large document processing workflows for information extraction and understanding its contents. Business intelligence processes like document retrieval, text recognition, content categorization, and others often require to extract the semantic information from documents when parsing the documents into a structured machine-readable format. This extracted data can be then integrated into document processing workflows in Robotic Process Automation tools. Thus, more efficient solutions have been developed in key industrial sectors (e.g. banking, finance, healthcare, and so on)~\cite{mathur2023layerdoc,pfitzmann2022doclaynet}. Document layout analysis (DLA) has become an important task in DI because any task related to document understanding entails the need of obtaining a structured representation that helps to localize the key information stored in them. Initially, remarkable progress has been observed with classical convolution-based algorithms (CNNs) such as Faster RCNN~\cite{sun2019faster} for Document Object Detection (DOD), Mask RCNN \cite{biswas2021beyond} for instance segmentation, among other specialized architectures. These architectures are quite simple to implement and effective in some specific case studies (e.g. table detection~\cite{li2020tablebank}, layout analysis of scientific articles~\cite{zhong2019publaynet} etc.) but they lack the generalization ability to address other similar tasks. Recently, Transformer-based architectures~\cite{appalaraju2021docformer,huang2022layoutlmv3} have achieved superior performance over CNNs with the help of a global attention mechanism. However, these models are not unified which prevents the mutual cooperation between the detection and segmentation tasks which affects their performance as the detection and segmentation modules cannot guide each other. Not only that, but those architectures were also biased toward their pre-trained datasets and failed to perform domain shifts for a similar task. As these transformer models are often pre-trained with massive amounts of data originating from a related source domain (i.e., large-scale industry documents~\cite{harley2015evaluation} or scientific articles~\cite{zhong2019publaynet}), they fail to address relatively different tasks (e.g. layout extraction in magazines~\cite{clausner2019icdar2019}). The introduction of this domain shift property to a DLA model has the potential to reduce computational expenses and help to create a more data-independent generic model.

To address the aforementioned issues, we propose \emph{SwinDocSegmenter} framework to perform instance-level segmentation of complex document layouts, using content query embeddings on a high-resolution pixel embedding map obtained from the Swin Transformer feature extraction   backbone~\cite{liu2021swin} and Transformer encoder features. It helps define global semantic reasoning of the features at a higher level which overcomes the drawbacks of using the ResNet-FPN~\cite{lin2017feature} backbone. Here, we initialize mask queries as anchors by utilizing the encoder dense prior to predicting the masks from the top-ranked tokens. It helps to perform pixel-wise segmentation at an early stage, which helps to enhance boxes. In the later stages, these boxes help to increase the segmentation performances by formulating dynamic anchor boxes. This phenomenon of mutual task cooperation helps to obtain a unified model for layout detection and segmentation. We introduce a contrastive denoising training inspired by \cite{zhang2022dino} to accelerate segmentation training by focusing on low-level instances. It boosts the model performance a lot as one of the main drawbacks of Transformers to working with unlabeled data where it penalizes the classes that have a very low number of feature representations \cite{biswas2022docsegtr}. Last but not the least, we utilize a hybrid bipartite matching \cite{li2022mask} for more consistent semantic matching which helps to perform \textit{domain shift} and utilizes the pre-trained weights of the transformers from a completely different domain to perform similar tasks. In this case, we utilized the pre-trained weights of the MS-COCO Object Detection benchmark~\cite{lin2014microsoft} for the instance segmentation  of complex document layouts.

The overall contributions of this work can be summarized in three folds:
\begin{itemize}
    \item A \textit{unified Transformer-based framework} has been proposed to perform instance-level document layout segmentation, with a Swin Transformer backbone, anchor box-guided cross-attention, and enhanced query selection strategy.
    \item We introduce \textit{contrastive denoising training} to enhance the low-level instances to boost the performance of the unlabeled dataset.
    \item We utilize \textit{hybrid bipartite matching to invoke the domain shift property} to save the pretraining time and use the publicly available pre-trained weights from diverse domains for a similar task which improves model generalization.
\end{itemize}

The rest of the paper is organized in the following way: In Section \ref{s:sota} we review state-of-the-art approaches for document layout analysis. We describe the \emph{SwinDocSegmenter} in Section \ref{s:method}. We introduce our experimental evaluation as well as ablation studies in Section \ref{s:experiments}. 
Finally, Section \ref{s:conclusion} draws the conclusion and guides the future research directions.

\section{Related Work}
\label{s:sota}
In order to extract the relevant information from digital documents, layout recognition methods obtain spatial understanding with relational reasoning between different layout components (e.g. table, text, figures, title, etc.). Mainstream layout analysis algorithms have been dominated by classical heuristic rule-based algorithms before the deep learning era. Later on, convolution frameworks play a leading role to solve this task until the transformers-based architectures achieve remarkable performance. This section is dedicated to obtaining an overview of the state-of-the-art for this task by analyzing different methodological schemes.

\paragraph{\textbf{Heuristic Rule-based Document Layout Analysis.}} Document layout segmentation using heuristic methods can be further classified into three different categories: top-down, bottom-up, and hybrid strategies. Bottom-up approaches \cite{asi2015simplifying,saabni2011language} perform basic operations like grouping and merging of pixels to create homogeneous regions for similar objects and separate them from the nonsimilar ones. Top-down strategies \cite{journet2005text,kise1998segmentation} split the document image into different regions iteratively, until a definite region has been obtained around similar objects. Although bottom-up approaches are able to tackle complex layouts, they are computationally expensive. Moreover, Top-down methods provide faster implementation but penalize the generalization, and perform effectively only on specific types of documents.  To take advantage of both, hybrid methods \cite{chen2011table,fang2011table} combine bottom-up and top-down cues to obtain fast and efficient results. Prior to the deep learning era, these methods were state-of-the-art for table detection.

\paragraph{\textbf{Convolution-based Document Layout Analysis.}} Since 2012, deep learning algorithms replaced the rule-based algorithm and Convolutional Neural Networks (CNNs) became the prior strategy to solve instance document segmentation tasks. Faster-RCNN \cite{ren2015faster} provides a strong document object detection that can be utilized to solve page segmentation \cite{li2020page}. Later on, a similar network Mask-RCNN \cite{almutairi2019instance} provides the first layout segmentation benchmark for instance segmentation of newspaper elements. Another convolution benchmark has been provided by RetinaNet \cite{lin2021keyword} for keyword detection in document images. This is a complex method and only helps to detect the text regions. In order to provide a new state-of-the-art benchmark for table detection and table structure recognition, DeepDeSRT \cite{schreiber2017deepdesrt} utilizes a novel image transformation strategy to identify the visual features of the table structures and feed them into a fully convolution network with skip pooling. Similarly, Oliviera et al \cite{oliveira2018dhsegment} used a similar FCNN-based framework for pixel-wise segmentation of historical document pages which outperforms the previous convolutional autoencoder-based benchmarks obtained by Chen et al \cite{chen2017convolutional,chen2015page}. Saha et al \cite{saha2019graphical} provided ICDAR2017 POD (Page Object Detection) benchmark \cite{gao2017icdar2017} to obtain state-of-the-art results by using transfer learning based Faster-RCNN backbone for detection of mathematical equations, tables, and figures. A new cross-domain DOD benchmark was established in \cite{Li_2020_CVPR} to apply domain adaptation strategies to solve the domain shift problem. Recently, A vision-based layout detection benchmark has been provided in \cite{yang2021vision} which utilized a recurrent convolutional neural network with VoVNet-v2 backbone \cite{lee2019energy} by generating synthetic PDF documents from ICDAR-2013 and GROTOAP dataset. It obtained a new benchmark to solve the scientific document segmentation task.

\paragraph{\textbf{Transformer Based Document Layout Analysis.}} Nowadays Transformers which provide a more prominent performance with the utilization of positional embedding and self-attention mechanism \cite{vaswani2017attention}. Here, DiT \cite{li2022dit} obtained a new baseline for document image classification, layout analysis, and table detection with self-supervised pretraining on large-scale unlabeled document images which cannot be applicable to small magazine datasets like PRIMA. Similarly, Li et al. \cite{li2021structext} obtained a multimodal framework to understand the structured text in the documents. However, the model performs very poorly for similar semantics of textual content. In order to improve these performances a TILT \cite{powalski2021going} mechanism has been introduced which simultaneously learns  textual semantics, visual features, and  layout information with an encoder-decoder Transformer. A similar transformer encoder-decoder was utilized in \cite{yang2022transformer} which provides a new baseline for the PubLayNet dataset (AP: 95.95) with the text information extracted through OCR. Recently, LayoutLmv3 \cite{huang2022layoutlmv3} used joint learning of text, layout, and visual features to obtain state-of-the-art results in visual document understanding (VDU) tasks. It performs significantly well for large-scale datasets but fails for small-scale datasets. DocSegTr \cite{biswas2022docsegtr} utilized a ResNet-FPN backbone over the transformer layers with self attention mechanism, which helps it to converge faster for small scale datasets but unable to achieve state-of-the-art performances. Other recent  approaches \cite{appalaraju2021docformer,kim2022ocr,gu2021unidoc,gu2022xylayoutlm} also utilize this joint pretraining strategy to solve several VDU tasks including document visual question answering. These techniques are quite helpful to several downstream tasks by a unified pretraining. However, it comes with a pretraining bias which prevents them to perform a domain shift and they also unable to learn the class information with low number of instances as their is no weight prioritizing.

Motivated by the recent breakthrough of transformers and to improve its performance by solving the above-mentioned issues we are proposing an end-to-end unified domain adaptive document segmentation transformer benefitted with contrastive training that not only achieves superior performance on standard instance-level segmentation benchmarks but also provides the first transformer baseline for the newly proposed industrial document layout analysis dataset \cite{pfitzmann2022doclaynet}.

\section{Method}
\label{s:method}
\begin{figure}[htb]
\vspace{-4mm}
\centering
  \includegraphics[width=\linewidth]{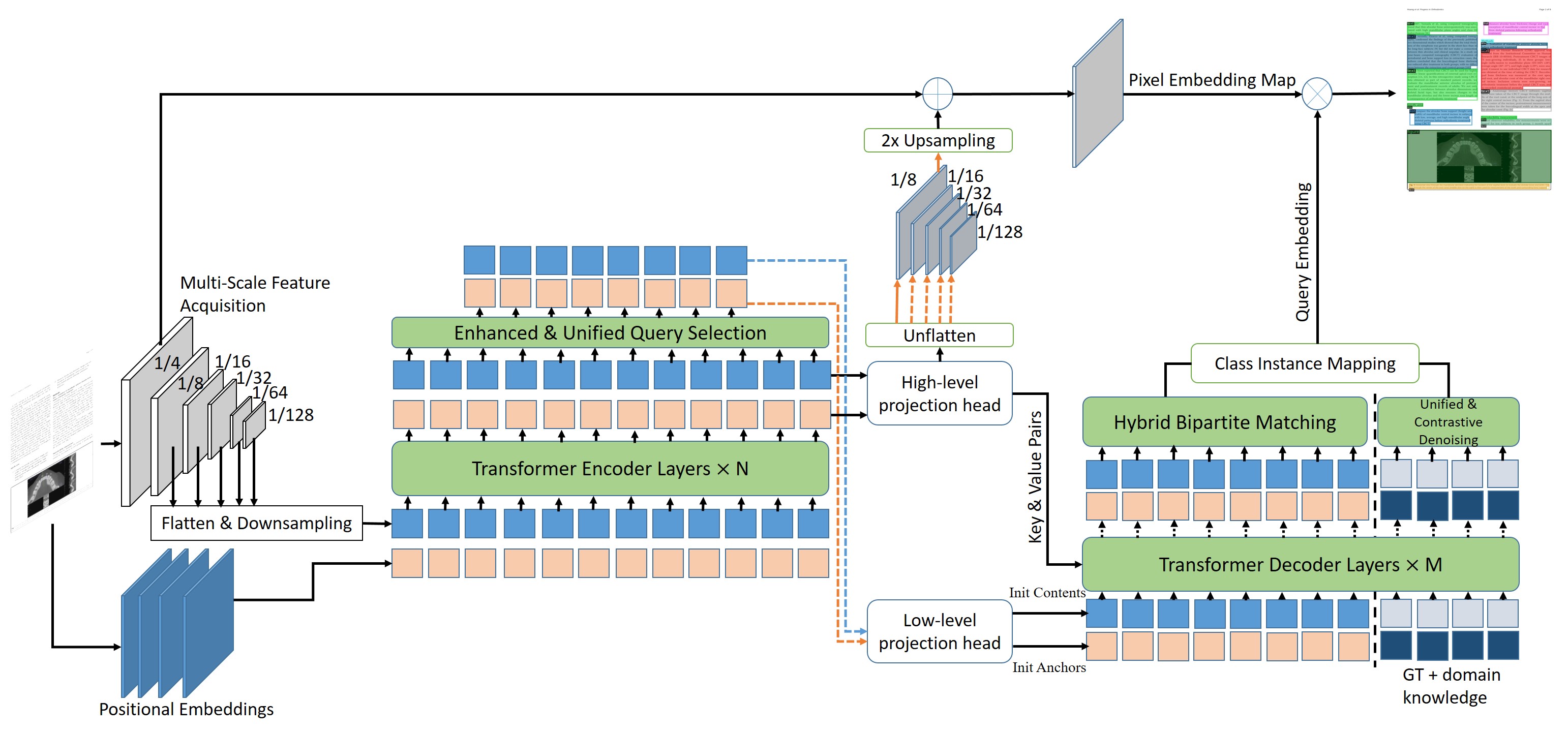}
\caption{\textbf{Proposed SwinDocSegmenter Framework.} Given an input document image from any domain, the model predicts the segmented document layout using a unified detection and segmentation branch}
\label{fig: method}       
\end{figure}

The proposed \emph{SwinDocSegmenter} is a unified end-to-end architecture that contains a Swin Transformer backbone \cite{liu2021swin}, a Transformer encoder-decoder pair, and a segmentation branch obtained from multiple projection heads by class instance mapping. The proposed architecture is illustrated in Fig. \ref{fig: method} where the model first extracts multi-scale features with a Swin backbone.
Then the features are flattened and downsampled before feeding them into the transformer encoder, otherwise it would generate a large number of trainable parameters which is impossible to train with limited resources. The Transformer encoder takes those features and their corresponding positional embeddings (obtained through several convolution layers with kernel size $3 \times 3$) as input to perform the feature enhancement. Here, a unified mixed query selection strategy has been obtained that passed through a low-level projection head to initialize the positional queries and anchors. The main advantage of this query selection strategy is that it does not initialize content queries but leaves them learnable which helps a lot in times of domain shift. Not only that, with the help of a low-level projection head it helps to focus on low-dimension image features which are often ignored in transformer training due to lack of data points. This also makes the decoder ready for contrastive denoising training (CDN) \cite{zhang2022dino}. In the decoder, deformable attention \cite{xia2022vision} is utilized to combine the outputs of the encoder with layer-by-layer query updates. With CDN it is also considered the segmented/wrongly segmented region as hard negative samples and tries to rectify it with a look forward twice approach \cite{zhang2022dino} where it passes the gradient between adjacent layers at early stages. We utilize a hybrid bipartite matching strategy to refine the segmented region based on the dynamic anchor boxes which help to generate an accurate segmented region. These two pieces of information are combined through class instance mapping to get the final query embedding. We perform a dot-product between the final query embedding and pixel embedding map to get the final instance segmentation output on document images.

\subsection{Segmentation Branch}
To perform mask classification, we utilize a key idea \cite{cheng2021mask2former} to construct a pixel embedding map (PEM) by combining the multi-scale features (extracted by Swin backbone) and Transformer encoded features. As shown in Fig. \ref{fig: method}, the PEM is constructed with a fusion between $1/4 ^ {th}$ resolution feature map from the backbone ($S_b$) and upsampled $1/8 ^ {th}$ resolution feature map from Transformer Encoder ($T_e$). The output mask $M$ is computed by a dot-product between PEM and query embedding ($Q_e$) obtained from the decoder (see eq. \ref{eq: 1}).

\begin{equation}
    M = Q_e \otimes \ethm (\Gamma(S_b) + \psi (T_e))
    \label{eq: 1}
\end{equation}

Where $\Gamma$ is the convolutional layer to map the channel dimension to the transformer dimension, $\psi$ is the interpolation function for $2 \times$ upsampling of $T_e$, and $\ethm$ is the segmentation head. This mechanism is simple and easy to implement.

\subsection{Feature encoding techniques}
The feature encoding techniques consist of four important subparts: Query selection, low-level and high-level feature projection, and anchor initialization to boost the performance and simplify the decoding technique. 

\paragraph{\textbf{Query selection strategy}}
It has been observed that the output of the encoder contains dense features that can be used as better priors in the decoder. Here, we adopted one classification, one detection, and one segmentation head, in the encoder output followed by a low-level and high-level projection head. We obtained the classification score of each token as a confidence score and used them to select top-ranked features and feed them into the decoder as content queries. The selected features also regress boxes via detection and segmentation heads and passed through the high-level projection head to combine with the high-resolution feature map via dot product to predict the masks. These predicted masks and boxes are considered initial anchors for the decoder after passing it through the low-level projection head. It helps to make the decoder for contrastive training as both high-level and low-level class instances are present and we can improve the performance of low-level class instances without compromising the performance of high-level instances by adjusting contrastive loss.

\paragraph{\textbf{Low-level projection head}}
It is a shallow multi-layer perceptron (MLP) that leverages the project features to low-level embeddings for contrastive learning in low-level views. It helps to learn more fine-grained invariances. Specifically, we apply a non-linear function $F = (f_1, f_2,..., f_s)$ on low-level features to enhance them before initializing them as content queries and anchors in decoders. The objective function of this low-level projection head is defined in eq. \ref{eq: 2}.
\begin{equation}
    \mathcal{L}_{low} = \sum_{i = 1}^{n} \sum_{j = 1}^{n'} -log \dfrac{exp(f_i \cdot f_j / \tau)}{\sum_{c=1}^{k} exp(f_c \cdot f_j/ \tau)}
    \label{eq: 2}
\end{equation}

Where, $n$ and $n'$ are the no. of features obtained from detection and segmentation heads, $c$ is the no. of top-ranked features obtained from the classification heads. Here, we need a temperature hyperparameter $\tau$ to tune the layers to enhance the features based on the datasets we have used. $\tau = 0.02, 0.6, 0.1,$ and $0.2$ for PublayNet, Prima, HJ, and TableBank respectively. Note: all these hyperparameter values have been obtained experimentally.

\paragraph{\textbf{High-level projection head}} It is a deep MLP that preserves the high-level invariance of the high-level features. Basically, it set a different number of prototypes $P = (p_1, p_2, ..., p_m)$ to obtained different key-value pairs $k_{1}v_{1}, k_{2}v_{2}, ..., k_{n}v_{n}$ which also enriched the feature representation. The objective function of this prototyping has been defined in eq. \ref{eq: 3}.
\begin{equation}
    \mathcal{L}_{high} = \sum_{i = 1}^{n} \sum_{j = 1}^{n'} -log \dfrac{exp(f_i \cdot p_j / \phi_{j})}{\sum_{c=1}^{k} exp(f_c \cdot p_j/ \phi_{j})}
    \label{eq: 3}
\end{equation}

Where, $p_j$ is the prototype of the corresponding key-value pairs and $\phi_j$ is the concentration estimation indicator \cite{mo2023multi} for the distribution of representations around the prototype.

\paragraph{\textbf{Anchor initialization}} Document instance segmentation is a classification task at the pixel level whereas, object detection is a position regression task at the region level. Therefore, segmentation is more challenging due to its fine granularity than detection though it is simpler to learn in the beginning. Dot-producting queries using the high-resolution feature map, for instance, can predict masks by only comparing semantic similarity per pixel. However, the box coordinates must be directly regressed for detection in an image. As a result, mask prediction is significantly more accurate than box prediction in the initial stage.  As a better anchor initialization for the decoder, therefore, we derive boxes from the predicted masks following unified query selection. The enhanced box initialization has the potential to significantly enhance the detection performance thanks to this efficient task cooperation.

\subsection{Feature decoding for mask prediction}
At this stage, we introduced unified contrastive denoising training for effectively boosting the performance for the low-level instances and hybrid bipartite matching to perform domain shift. Below, we discuss both strategies in detail.

\begin{figure}
\vspace{-4mm}
\centering
  \includegraphics[width=\linewidth]{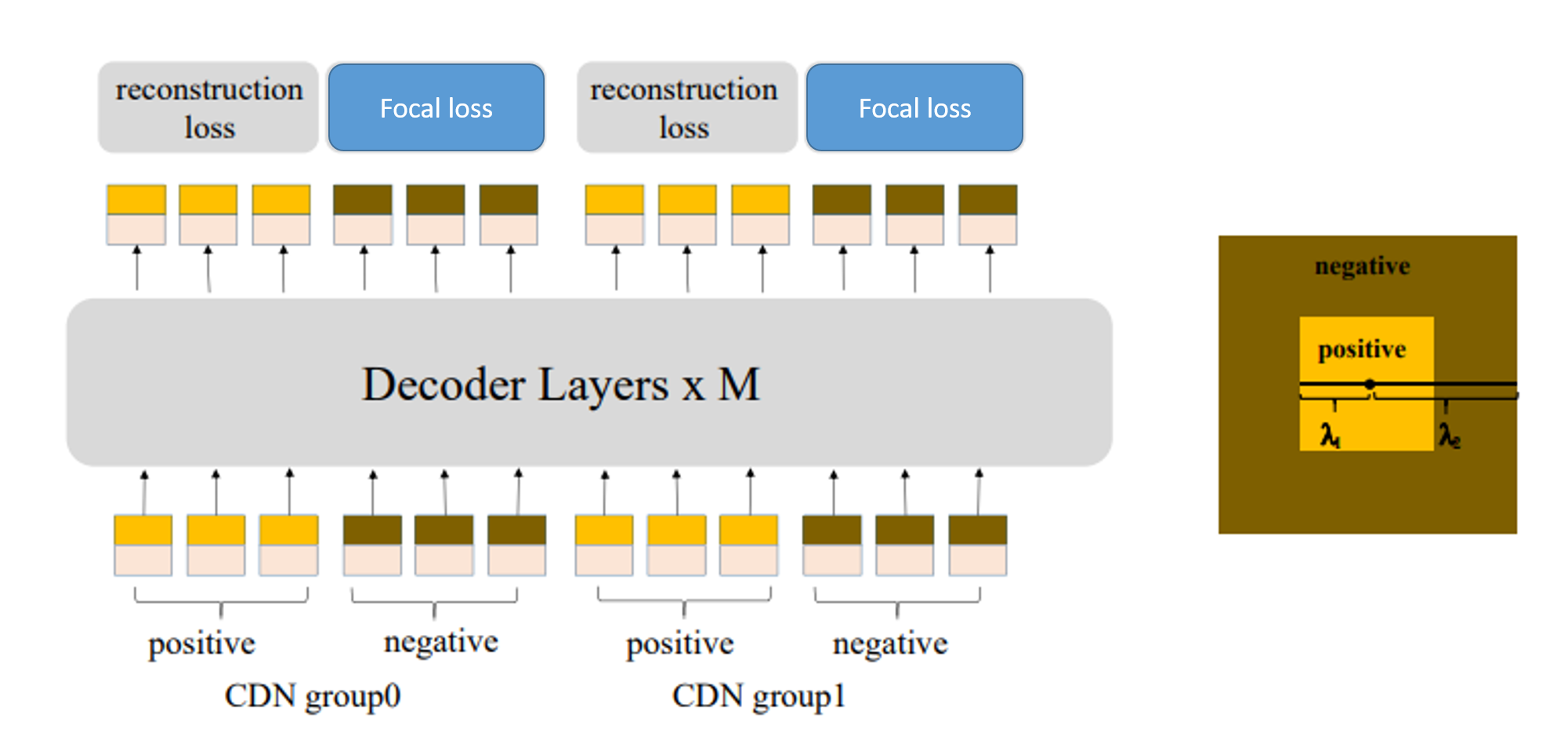}
\vspace{1mm}
\caption{\textbf{Unified Contrastive Denoising Training Strategy.} Similar to DINO \cite{zhang2022dino} implementation, however, in place of no object detection we introduced a focal loss to optimize and enhance the low-level instances.}
\label{fig: decoder}       
\end{figure}

\paragraph{\textbf{Unified Contrastive Denoising Training}} Query denoising \cite{zhang2022dino} is an effective technique to improve performance by accelerating convergence. However, it lacks the capability of separating two nearby class instances. CDN can tackle this issue by rejecting useless anchors. Here, noises are added to ground truth labels and boxes, and the Transformer decoder receives them as noised positional and content queries. Here, we have two hyperparameters $\lambda_p$ and $\lambda_e$ where, $\lambda_e >  \lambda_p$ as depicted in Fig. \ref{fig: decoder}.
It helps to generate two types of queries (positive and negative). It is anticipated that positive queries within the inner square will reconstruct their corresponding ground truth boxes because their noise scale is less than $\lambda_p$. On the other hand, negative queries have a noise scale greater than $\lambda_p$ and less than $\lambda_e$ which are minimized through the focal loss. Generally, we keep $\lambda_e$ very small as it helps to improve the performance by keeping the hard negative samples close to the ground truth anchors. Each CDN group has positive and negative queries (see Fig. \ref{fig: decoder}). A CDN group will have $2 \times q$  queries for an image with $q$ GT boxes, with each GT box producing a positive and negative query. To increase the efficiency we also employ multiple CDN groups.

In order to train the model, the noised versions of the object features have been utilized to reconstruct them. We also apply this method to tasks involving segmentation. Boxes and masks are naturally connected due to the fact that masks can be seen as a more finely detailed representation of boxes. As a result, we can train the model to predict masks given boxes as a denoising task and treat boxes as a noised version of masks. In order to train mask denoising more effectively, the boxes provided for mask prediction are also randomly noised. During training, these noised objects will be added to the original decoder queries, but they will be removed during inference. We perform a lot of tunning to get the optimized value of $\lambda_p$ and $\lambda_e$. However,  it has been observed that, most generic performance has been achieved with $\lambda_p = 0.1$ and $\lambda_e = 0.02$ respectively.

\paragraph{\textbf{Hybrid bipartite matching}}
This technique helps to remove the inconsistency between the pair of masks predicted from different heads by changing their corresponding weights. With this motivation, we utilize this concept in domain shift. Basically, we add an extra mask prediction loss in addition to the L1 and focal loss in bipartite matching. It encourages more accurate and consistent matching results for one query. So, when we utilized a pre-trained model from a different domain we penalize this loss more which forced us to make significant changes in their corresponding weights and slowly decrease the penalizing rate when it reached near the convergence. This loss is also optimized along with the L1 and focal loss to make this domain shift unified.
Finally, a class instance mapping is performed between the classes and the predicted instances. It is a simple one-to-one mapping to perform query embedding which can be combined with pixel embedding map effectively through dot product in order to complete the instance segmentation process in and end-to-end manner.

\section{Experimental Evaluation}
\label{s:experiments}

\textbf{Datasets.} The Document Layout Analysis (DLA) community has always been concerned about the absence of standard public benchmarks. We use large-scale annotated datasets like PubLayNet \cite{zhong2019publaynet}, TableBank \cite{prasad2020cascadetabnet}, and Historical Japanese (HJ) \cite{shen2020large} as well as small-scale PRIMA \cite{clausner2019icdar2019} for evaluating our proposed segmentation approach in this work (Please refer to Table \ref{tab:01} for a detailed description). Besides that, we evaluate our model against a recently released standard industrial document layout segmentation benchmark~\textbf{DocLayNet}~\cite{pfitzmann2022doclaynet}. It contains 91104 object instances of 11 distinct labels (Caption, Footnote, Formula, List-item, Page-footer, Page-header, Picture, Section-header, Table, Text, and Title) and covers a wide range of document object sizes (large to small).\\ 

\begin{table}[h]
\caption{Experimental dataset description (instance level)}
\label{tab:01}
\centering
\resizebox{\linewidth}{!}{
\begin{tabular}{@{}cccccccccccc@{}}
\toprule
\multicolumn{3}{c}{\textbf{PublayNet}}           & \multicolumn{3}{c}{\textbf{PRIMA}}               & \multicolumn{3}{c}{\textbf{Historical Japanese}} & \multicolumn{3}{c}{\textbf{TableBank}}           \\ \midrule
\textbf{Object} & \textbf{Train} & \textbf{Eval} & \textbf{Object} & \textbf{Train} & \textbf{Eval} & \textbf{Object} & \textbf{Train} & \textbf{Eval} & \textbf{Object} & \textbf{Train} & \textbf{Eval} \\ \midrule
Text            & 2,343,356      & 88,625        & Text            & 6401           & 1531          & Body            & 1443           & 308           & Table           & 2835           & 1418          \\ \midrule
Title           & 627,125        & 18,801        & Image           & 761            & 163           & Row             & 7742           & 1538          & -               & -              & -             \\ \midrule
Lists           & 80,759         & 4239          & Table           & 37             & 10            & Title           & 33,637         & 7271          & -               & -              & -             \\ \midrule
Figures         & 109,292        & 4327          & Math            & 35             & 7             & Bio             & 38,034         & 8207          & -               & -              & -             \\ \midrule
Tables          & 102,514        & 4769          & Separator       & 748            & 155           & Name            & 66,515         & 7257          & -               & -              & -             \\ \midrule
-               & -              & -             & other           & 86             & 25            & Position        & 33,576         & 7256          & -               & -              & -             \\ \midrule
-               & -              & -             & -               & -              & -             & Other           & 103            & 29            & -               & -              & -             \\ \midrule
\textbf{Total}  & \textbf{3,263,046}      & \textbf{120,761}       & \textbf{Total}  & \textbf{8068}           & \textbf{1891}          & \textbf{Total}  & 181,097        & 31,866        & \textbf{Total}  & 2835           & 1418          \\ \bottomrule
\end{tabular}
}
\vspace{-4mm}
\end{table}


\noindent
\textbf{Evaluation Metrics.} The Intersection over Union (IoU) score is the most general way to assess the accuracy of the predicted instance (document category) for an instance-level segmentation task. Standard Microsoft COCO benchmark evaluation for instance segmentation uses the mean of APs at various IoU thresholds (0.5 to 0.95 with a step size of 0.05) to calculate the mean Average Precision (mAP) score for the entire model. Since all of them use a similar environment to compute the mAP, comparing the proposed approach to those that are already in use is helpful. In addition, the model performance for evaluating each categorical document instance has been calculated in accordance with \cite{biswas2022docsegtr,huang2022layoutlmv3,shen2021layoutparser}.\\

\noindent
\textbf{The Choice of the Feature Extraction Backbone.} In the context of instance-level document segmentation, extensive ablation studies were carried out to quantify the significance of each component of our model framework and to justify its use for segmenting various layout elements. All the ablations have been performed on the PRIMA dataset as it is the smallest dataset and it contains difficult layouts. In this study, different CNN and Vision Transformer backbones has been used (see Table \ref{tab:1}). Among them, we take the SwinL Transformer backbone to multi-scale feature extraction. Though the no. of trainable parameters increases,  it also improves the performance over ResNet, ResNeXt, and ViTs by $~8\%$, and from Swin Tiny by $~5\%$. The convolutional backbones provide attention to local features which are effective for small object detection however, there is no global attention that penalizes the cost for the large object. On the other hand, ViTs utilize self-attention but require a large amount of training data to learn the multi-scale features. Initially, Swin-Tiny will perform well but it is sensitive to noise so at a later stage, it penalizes the reconstruction which affects the overall performance. Due to its large size SwinL can eliminate noise very easily and achieves better performance than the rest.\\

\begin{table}[h]
\vspace{-10mm}
\centering
\caption{Ablation Study of different feature extraction backbones}
\label{tab:1}
\begin{tabular}{@{}cccccccc@{}}
\toprule
Backbone    & No. of Parameters & AP     & AP@50  & AP@75  & APs & APm & APl \\ \midrule
ResNet-50   & 52M               & 36.065 & 52.362 & 41.112 & 20.152       & 23.327       & 38.142       \\
ResNet-101  & 102M              & 37.112 & 54.982 & 41.872 & 22.242       & 26.153       & 41.986       \\
ResNext-101 & 104M              & 38.405 & 58.405 & 41.916 & 25.982       & 29.364       & 44.129       \\ \midrule
ViT-S       & 126M              & 40.342 & 59.763 & 42.158 & 29.176       & 33.129       & 48.526       \\
ViT-B       & 164M              & 46.128 & 62.689 & 47.358 & 31.389       & 33.458       & 50.508       \\ \midrule
Swin-T      & 178M              & 49.349 & 65.956 & 50.317 & 34.128       & 36.909       & 52.049       \\
Swin-L      & 223M              & \textbf{54.393} & \textbf{69.313} & \textbf{52.965} & \textbf{39.327}       & \textbf{42.061}       & \textbf{60.142}       \\ \bottomrule
\end{tabular}
\end{table}

\noindent
\textbf{The Choice of the Input Image Resolution.} Besides that, the image resolution also affects the model performance as the model is large and the number of trainable parameters is huge. So if we use the small image resolution then at the late stage, it only learns the noise which wastes the computational resources. Increasing the image resolution improves the performance (see Table \ref{tab:2}), however, we are unable to increase beyond 1024 due to the limited resources. Moreover, from the trend, it can be concluded that increasing image resolution also improves the system performance until it meets the saturation point.\\

\begin{table}[h]
\vspace{-12mm}
\centering
\caption{How image resolution affects the instance segmentation performance}
\label{tab:2}
\begin{tabular}{@{}ccccccc@{}}
\toprule
Image Resolution & AP     & AP@50  & AP@0.75 & APs    & APm    & APl    \\ \midrule
$256 \times 256$        & 45.022 & 60.189 & 46.258  & 28.372 & 32.458 & 53.568 \\
$512 \times 512$        & 50.132 & 66.235 & 52.317  & 32.242 & 36.909 & 54.148 \\
$1024 \times 1024$      & 54.393 & 69.313 & 52.965  & 39.327 & 42.061 & 60.142 \\ \bottomrule
\end{tabular}
\end{table}

\noindent
\textbf{The Choice of the number of Decoder Queries.} 
Similarly, by taking a deep dive into the model we observe that, the no. of queries used for initializations in the decoder affects the overall performance.  With a small number of queries it will be very difficult to generate the negative samples close to the performance, which not only penalizes the model performance but also increases the optimization time of the loss function. Also, dense queries stabilize the model and  provide an opportunity to rectify the misclassified samples (see Table \ref{tab:3}). With the SwinL backbone, we can extend it to 900-1200 but we have to restrict it to 300 due to the limited computational resources.\\

\begin{table}[h]
\vspace{-10mm}
\centering
\caption{Ablation Study on No. of queries generated from Transformer Encoder}
\label{tab:3}
\begin{tabular}{@{}ccccccc@{}}
\toprule
\textbf{No. of Queries} & \textbf{AP}     & \textbf{AP@50}  & \textbf{AP@0.75} & \textbf{APs}    & \textbf{APm}    & \textbf{APl}    \\ \midrule
100 & 50.022 & 65.189 & 52.258 & 32.372 & 36.458 & 53.968 \\
150 & 50.132 & 66.235 & 52.317 & 32.242 & 36.909 & 54.148 \\
200 & 51.393 & 67.313 & 52.765 & 37.312 & 41.011 & 60.111 \\
250                     & 52.092          & 68.212          & 52.964  & 37.512          & 42.060 & 60.132 \\
300                     & \textbf{54.393} & \textbf{69.313} & \textbf{52.965}  & \textbf{39.327} & \textbf{42.061} & \textbf{60.142} \\ \bottomrule
\end{tabular}
\end{table}

\begin{figure}[h]
\centering
  \includegraphics[width=3in]{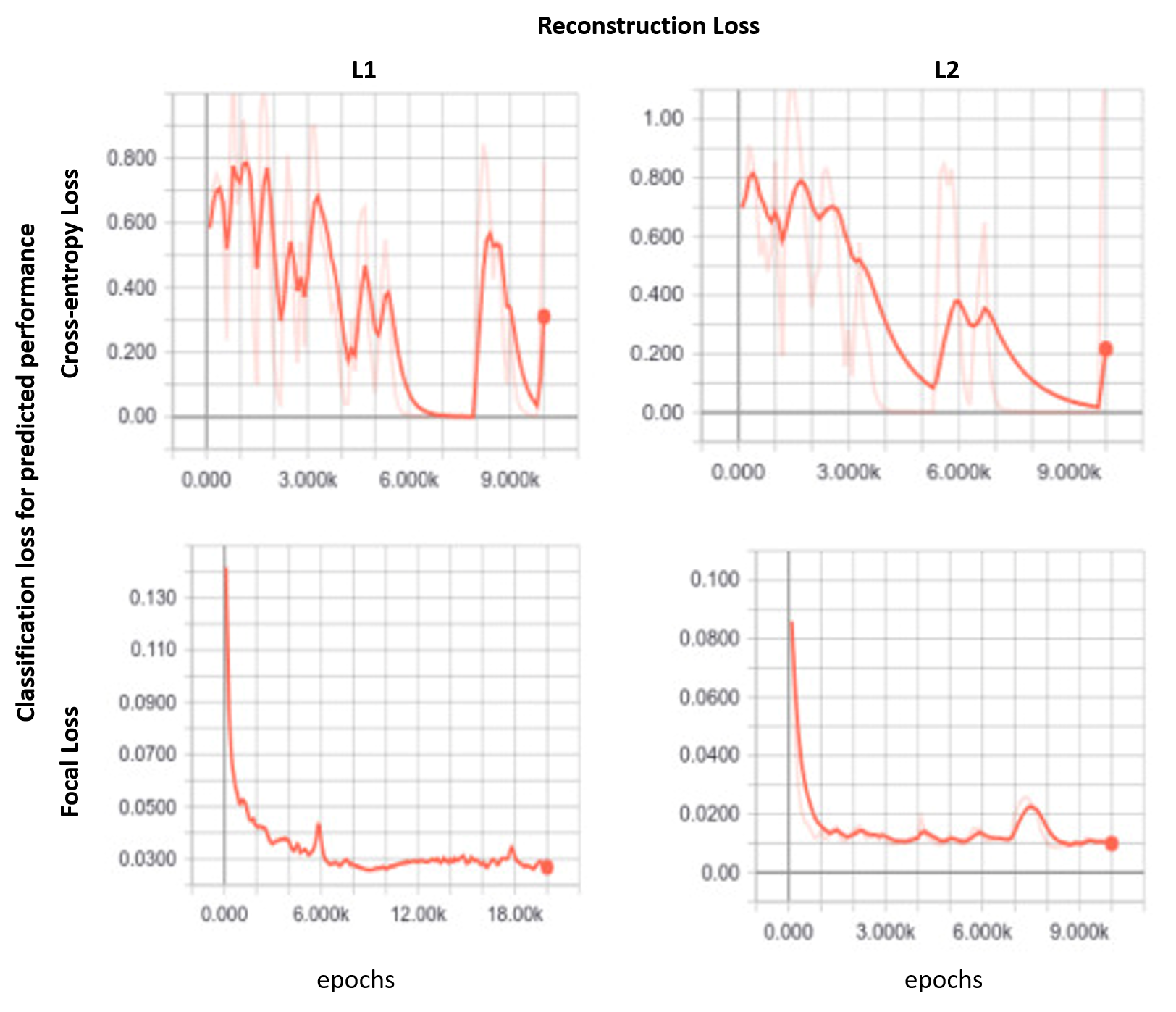}
\caption{\textbf{The impact of learning objectives.} The above graphs study the effectiveness of different combinations of loss functions}
\label{fig: loss}       
\end{figure}

\noindent
\textbf{The Choice of the Learning Objectives.} Last, but not the least an ablation study of the loss functions has been obtained to understand which combination of the reconstruction and classification loss is most optimized. From Fig.\ref{fig: loss} it has been observed that the combination of L1 and focal loss is the most effective one for this task. The L1 loss tends to shrink coefficients to zero which is better for feature selection whereas L2 tends to shrink coefficients evenly. On the other hand, Focal Loss helps to scale the standard cross-entropy loss to down-weight loss corresponding to easily classifiable examples dynamically and focus more on hard examples to make the system perform better on hard examples as well.\\

Moreover, the most interesting fact has been observed in Table \ref{tab: 07} which shows how pre-training on similar dataset include biases and shrinks the overall performance of the model. Here, we have used one SwinL backbone pre-trained on the PubLayNet dataset and another one pre-trained on the MSCOCO dataset. We can observe that the model achieves very high performance on the class "Table" because it 
is a common class in both datasets. But as the pre-trained model is not familiar with the "Separator" region, it penalizes a lot decreasing the overall performance of the networks as it quickly converges the loss by looking at the similar classes and ignoring the others by taking them as negative samples in CDN. Whereas with the MSCOCO pre-training it achieves a generic performance due to the enhanced and unified query selection which let the content queries learnable, and hybrid bipartite matching helps to optimize the corresponding pre-training weights. This helps to eliminate the bias factor. \\
\begin{table}[h]
\vspace{-10mm}
\centering
\caption{How pre-training provides biases}
\label{tab: 07}
\resizebox{\linewidth}{!}{
\begin{tabular}{@{}c|cccccc|cccccc@{}}
\toprule
\multirow{2}{*}{pre-training} & \multicolumn{6}{c}{Overall Performance}                & \multicolumn{6}{c}{Class-Wise Performance}                                       \\ \cmidrule(l){2-13} 
                             & AP             & AP@50 & AP@75 & APs   & APm   & APl   & Text  & Image & Table          & Math           & Separator      & Other         \\ \midrule
PubLayNet                    & 49.36          & 64.43 & 51.45 & 32.94 & 34.07 & 54.21 & 85.55 & 72.51 & \textbf{70.68} & 56.05          & 8.55           & 2.83          \\
Ms-COCO                      & \textbf{54.39} & 69.31 & 52.96 & 39.32 & 42.06 & 60.14 & 87.72 & 75.92 & 49.89          & \textbf{78.19} & \textbf{27.56} & \textbf{7.05} \\ \bottomrule
\end{tabular}
}
\end{table}

\begin{figure*}[h]
\vspace{-4mm}
\subfloat[LayoutParser]{%
  \includegraphics[width=0.5\linewidth,height=4cm]{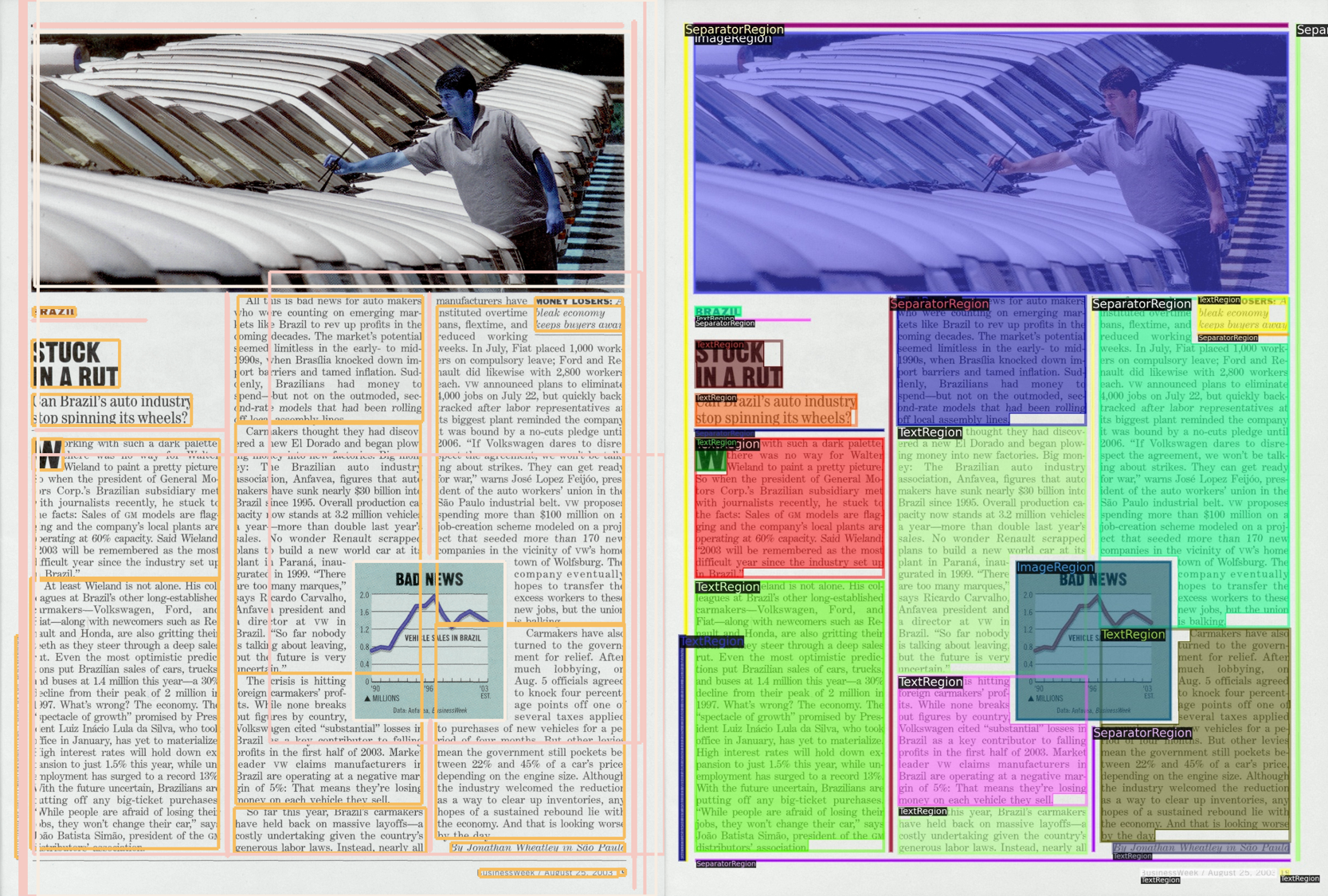}%
}\quad
\subfloat[LayoutLMv3]{%
  \includegraphics[width=0.5\linewidth,height=4cm]{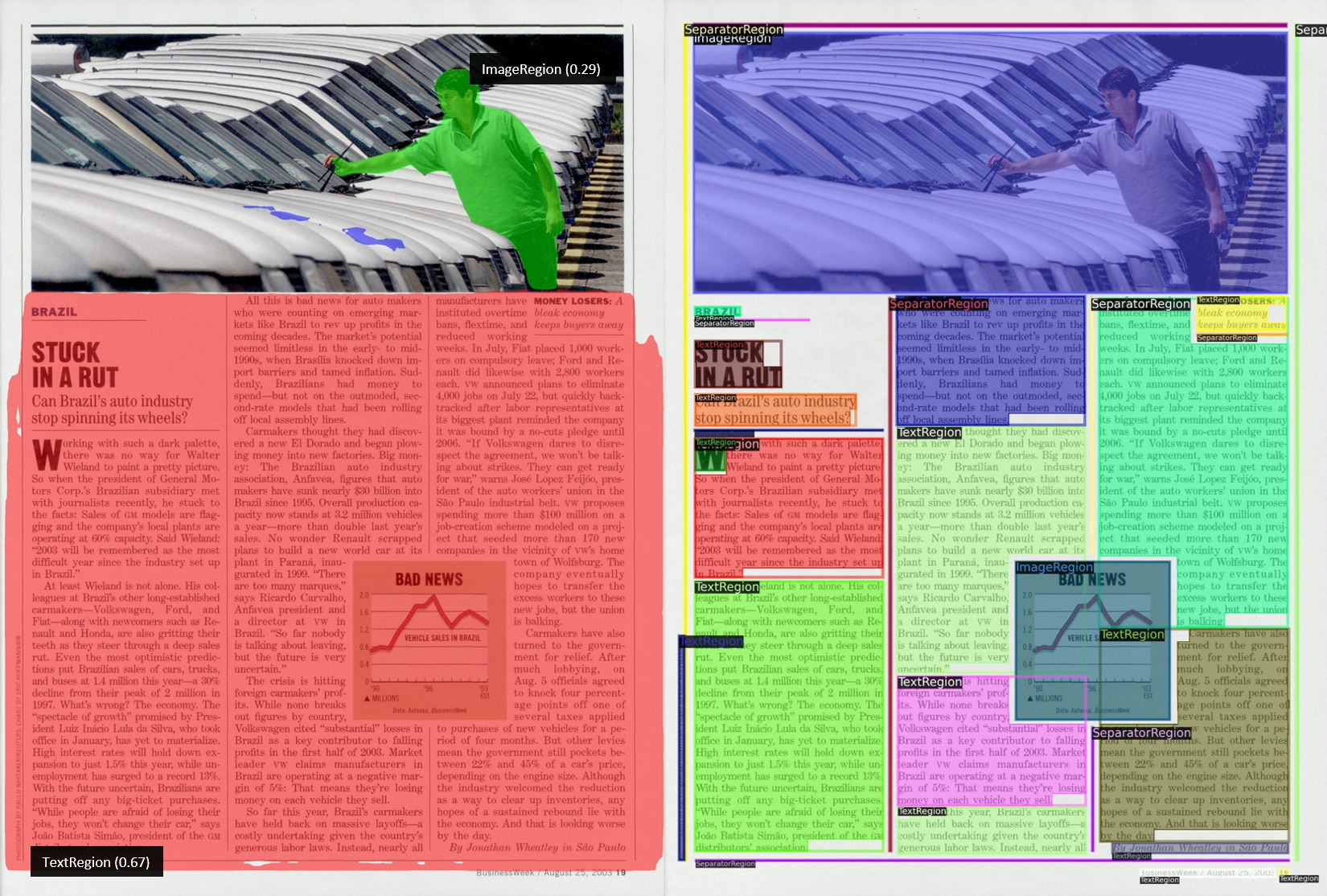}%
}\\
\subfloat[DocSegTr]{%
  \includegraphics[width=0.5\linewidth,height=4cm]{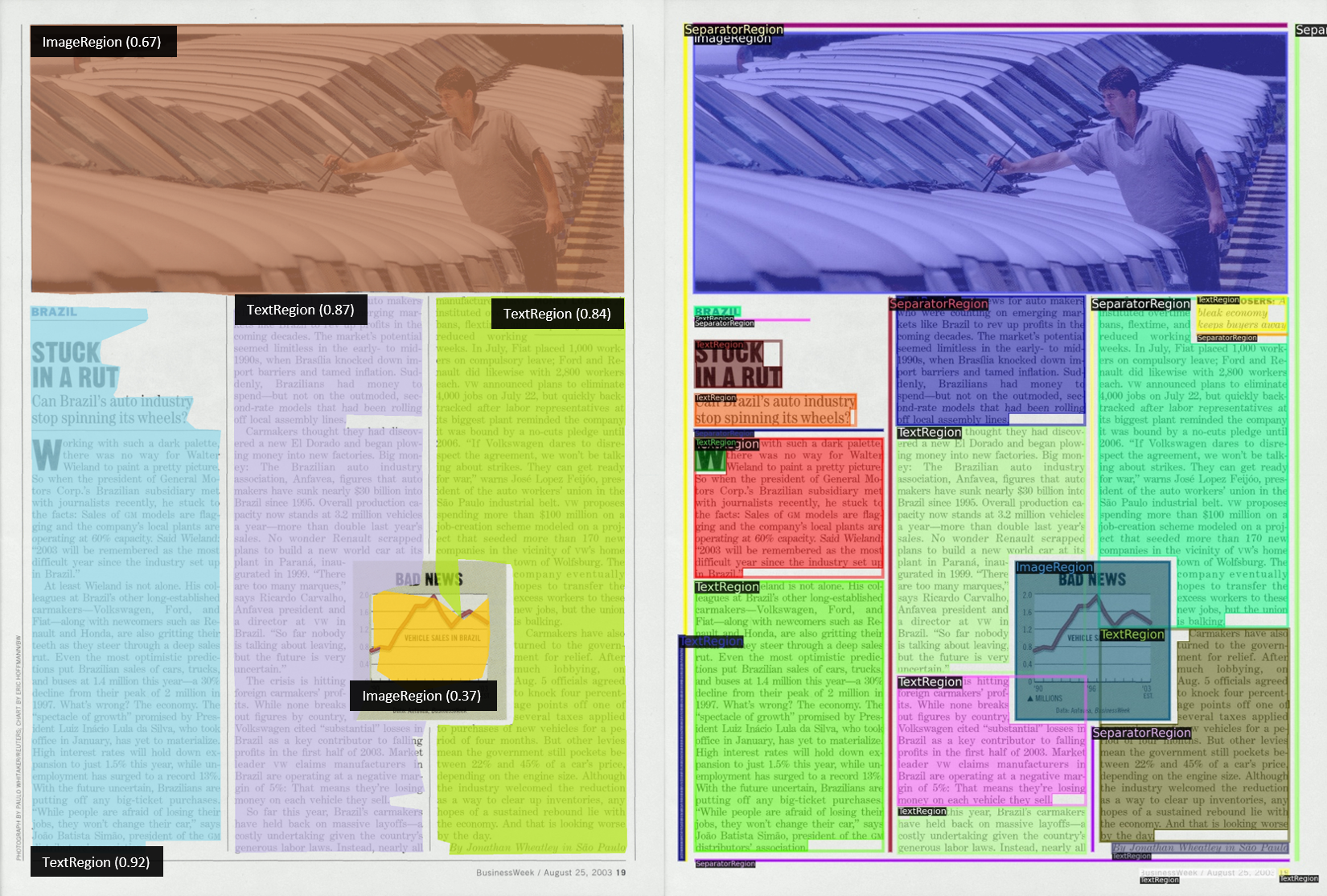}\quad
}
\subfloat[Ours]{%
  \includegraphics[width=0.5\linewidth,height=4cm]{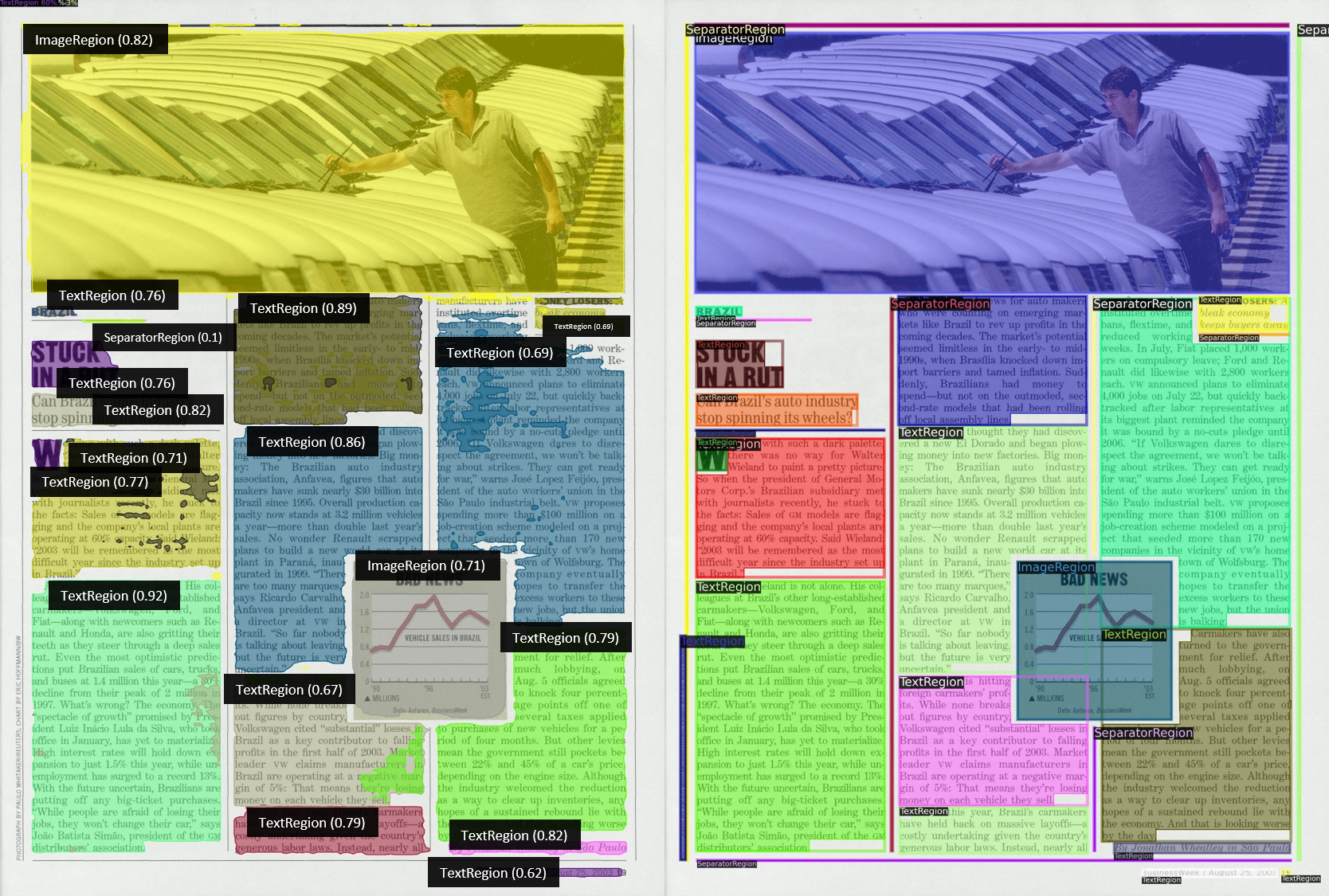}
}
\caption{Comparative analysis of the SwinDocSegmenter framework with the state-of-the-art approaches (\textbf{Left:} Predicted layout \textbf{Right:} Ground-truth)}
\label{fig:examples_detection_all}
\end{figure*}

\noindent
\textbf{Qualitative Insights.} 
The layout segmentation results on the PRIMA dataset obtained by \emph{SwinDocSegmenter} and state-of-the-art approaches are shown in Fig. \ref{fig:examples_detection_all}. In this test case, \emph{SwinDocSegmenter} is able to segment instances of different layout elements quite effectively.

It can be observed from Fig. \ref{fig:examples_detection_all}(a), though LayoutParser is quite effective for PRIMA dataset, it fails for this complex case as the bounding boxes are quite overlapping and not properly mapped with the ground truth. However, LayoutLMv3 (Fig. \ref{fig:examples_detection_all}(b)) performs very poor in this case. It identifies text and figure instances but not the other class instances. DocSegTr (Fig. \ref{fig:examples_detection_all}(c)) tries to improve the performance but it is still far away from the ground truth segmentation. On the other hand, our method segments complex document more satisfactorily and also maps the class instances with the ground truth (Fig. \ref{fig:examples_detection_all}(d)).\\

\begin{table*}[h]
\centering
\caption{Performance on DocLayNet Benchmark}
\label{tab:doclay}
\begin{tabular}{@{}ccccc@{}}
\toprule
Classes        & MaskRCNN & FasterRCNN & Yolov5        & Ours  \\ \midrule
Caption        & 71.5     & 70.1       & 77.7          & \textbf{83.56} \\
Footnote       & 71.8     & 73.7       & \textbf{77.2}          & 64.82 \\
Formula        & 63.4     & 63.5       & \textbf{66.2}          & 62.31 \\
List-item      & 80.8     & 81.0       & \textbf{86.2}          & 82.33 \\
Page-footer    & 59.3     & 58.9       & 61.1          & \textbf{65.11} \\
Page-header    & 70.0     & \textbf{72.0}       & 67.9          & 66.35 \\
Picture        & 72.7     & 72.0       & 77.1          & \textbf{84.71} \\
Section-header & 69.3     & 68.4       & \textbf{74.6}          & 66.5  \\
Table          & 82.9     & 82.2       & \textbf{86.3}          & \textbf{87.42} \\
Text           & 85.8     & 85.4       & 88.1          & \textbf{88.23} \\
Title          & 80.4     & 79.9       & \textbf{82.7}          & 63.27 \\ \midrule
All            & 73.5     & 73.4       & \textbf{76.8} & \textbf{76.85} \\ \bottomrule
\end{tabular}
\vspace{-4mm}
\end{table*}

\noindent
\textbf{Quantitative Analysis.} The final performance of the \emph{SwinDocSegmenter} in terms of mAP is quite interesting and it has the ability to provide a new benchmark for Document Layout Segmentation. In Table \ref{tab:doclay} and \ref{tab:04}, the method achieves a second position as both the LayoutLMv3 \cite{huang2022layoutlmv3} and Layout Parser \cite{shen2021layoutparser} use text information along with the visual information for instance segmentation task. In the case of PublayNet, we observe that the proposed is better identifying the text region than LayoutLMv3 and it provides comparable performance for other categories except for the "Title". Now "Title" also contains text so without textual information, it will be very difficult to solve these borderline cases. Also in terms of $AP@0.5$ and $AP@0.75$, it already surpasses LayoutLMv3 by only using visual information. Not only that, but it also outperforms the DiT (AP: 93.5) \cite{li2022dit} and achieves comparable performance with UDoc (AP: 93.9) \cite{gu2021unidoc} which provide a standard benchmark on the PubLayNet dataset. The same observations have been noticed for the PRIMA dataset. It is already observed that LayoutLMv3 is not good enough to detect small objects. But Layout Parser can, as it has a convolution backbone instead of a Transformer backbone and it also uses Microsoft OCR to extract the textual information from the images to combine them with visual information. The proposed model surpasses this state-of-the-art for all categories except "Table" and "Others". The performance is mainly affected by the "others" category as there is no such particular definition and without proper text information it is very difficult to separate them from the other categories.

\begin{table*}[h]
\vspace{-8mm}
\centering
\caption{Performance Analysis on the PubLayNet and PRIMA Benchmark}
\label{tab:04}
\begin{tabular}{@{}cccccccccc@{}}
\toprule
\multicolumn{5}{c}{\textbf{PublayNet}}                          & \multicolumn{5}{c}{\textbf{PRIMA}}                       \\ \midrule
\textbf{Object} &
  \textbf{\begin{tabular}[c]{@{}c@{}}Layout\\ Parser\end{tabular}} &
  \textbf{\begin{tabular}[c]{@{}c@{}}Doc\\ SegTr\end{tabular}} &
  \textbf{\begin{tabular}[c]{@{}c@{}}Layout\\ LMv3\end{tabular}} &
  \textbf{Ours} &
  \textbf{object} &
  \textbf{\begin{tabular}[c]{@{}c@{}}Layout\\ Parser\end{tabular}} &
  \textbf{\begin{tabular}[c]{@{}c@{}}Doc\\ SegTr\end{tabular}} &
  \textbf{\begin{tabular}[c]{@{}c@{}}Layout\\ LMv3\end{tabular}} &
  \textbf{Ours} \\ \midrule
\textbf{Text}    & 90.1 & 91.1 & 94.5          & \textbf{94.55} & \textbf{Text}      & 83.1 & 75.2 & 70.8 & \textbf{87.72} \\
\textbf{Title}   & 78.7 & 75.6 & 90.6          & 87.15          & \textbf{Image}     & 73.6 & 64.3 & 50.1 & \textbf{75.92} \\
\textbf{Lists}   & 75.7 & 91.5 & 95.5          & 93.03          & \textbf{Table}     & 95.4 & 59.4 & 42.5 & 49.89          \\
\textbf{Figures} & 95.9 & 97.9 & 97.9          & 97.91          & \textbf{Math}      & 75.6 & 48.4 & 46.5 & \textbf{78.19} \\
\textbf{Tables}  & 92.8 & 97.1 & 97.9          & 97.25          & \textbf{Separator} & 20.6 & 1.8  & 9.6  & \textbf{27.56} \\
\textbf{}        &      &      &               &                & \textbf{other}     & 39.7 & 3.0  & 17.4 & 7.054          \\ \midrule
\textbf{AP}      & 86.7 & 90.4 & \textbf{95.1} & 93.72          & \textbf{AP}        & 64.7 & 42.5 & 40.3 & 54.39          \\
\textbf{AP@0.5}  & 97.2 & 97.9 &               & \textbf{97.94} & \textbf{AP@0.5}    & 77.6 & 54.2 &      & 69.31          \\
\textbf{AP@0.75} & 93.8 & 95.8 &               & \textbf{96.28} & \textbf{AP@0.75}   & 71.6 & 45.8 &      & 52.965         \\ \bottomrule
\end{tabular}
\end{table*}

In Table \ref{tab:05} it has been observed that the proposed method outperforms all the previous state-of-the-art approaches. It outperforms the DocSegTr in the Historical Japanese dataset by a small margin ($~1\%$) but shows a significant improvement in the "Name" and "Position" categories. On the other hand, it shows a significant improvement ($~5\%$) in the Table Detection task on the TableBank dataset as it has only one category and comparatively less challenging layouts.

\begin{table*}[h]
\vspace{-6mm}
\centering
\caption{Performance Analysis on HJ and TableBank Benchmark}
\label{tab:05}
\begin{tabular}{@{}cccccccccc@{}}
\toprule
\multicolumn{5}{c}{\textbf{Historical Japanese}}        & \multicolumn{5}{c}{\textbf{TableBank}}                 \\ \midrule
\textbf{object} &
  \textbf{\begin{tabular}[c]{@{}c@{}}Layout\\ Parser\end{tabular}} &
  \textbf{\begin{tabular}[c]{@{}c@{}}Doc\\ SegTr\end{tabular}} &
  \textbf{\begin{tabular}[c]{@{}c@{}}Layout\\ LMv3\end{tabular}} &
  \textbf{Ours} &
  \textbf{object} &
  \textbf{\begin{tabular}[c]{@{}c@{}}Layout\\ Parser\end{tabular}} &
  \textbf{\begin{tabular}[c]{@{}c@{}}Doc\\ SegTr\end{tabular}} &
  \textbf{\begin{tabular}[c]{@{}c@{}}Layout\\ LMv3\end{tabular}} &
  \textbf{Ours} \\ \midrule
\textbf{Body}     & 99.0 & 99.0 & 99.0 & \textbf{99.72} & \textbf{Table}   & 91.2 & 93.3 & 92.9 & \textbf{98.04} \\
\textbf{Row}      & 98.8 & 99.1 & 99.0 & 99.0           & \textbf{}        &      &      &      &                \\
\textbf{Title}    & 87.6 & 93.2 & 92.9 & 89.5           & \textbf{}        &      &      &      &                \\
\textbf{Bio}      & 94.5 & 94.7 & 94.7 & 86.26          & \textbf{}        &      &      &      &                \\
\textbf{Name}     & 65.9 & 70.3 & 67.9 & \textbf{83.8}  & \textbf{}        &      &      &      &                \\
\textbf{Position} & 84.1 & 87.4 & 87.8 & \textbf{93.0}  & \textbf{}        &      &      &      &                \\
\textbf{Other}    & 44.0 & 43.7 & 38.7 & 40.57          & \textbf{}        &      &      &      &                \\ \midrule
\textbf{AP}       & 81.6 & 83.1 & 82.7 & \textbf{84.55} & \textbf{AP}      & 91.2 & 93.3 & 92.9 & \textbf{98.04} \\
\textbf{AP@0.5}   &      & 90.1 &      & \textbf{90.78} & \textbf{AP@0.5}  &      & 98.5 &      & \textbf{98.95} \\
\textbf{AP@0.75}  &      & 88.1 &      & \textbf{88.22} & \textbf{AP@0.75} &      & 94.9 &      & \textbf{98.90} \\ \bottomrule
\end{tabular}
\end{table*}

Last but not the least, we have obtained the first Transformer based baseline for a newly proposed dataset \textbf{DocLayNet} which contains industrial documents and the layouts are more challenging than PubLayNet benchmark. From Table \ref{tab:doclay} we conclude that our proposed SwinDocSegmenter outperforms the convolutional-based algorithms (MaskRCNN, FasterRCNN, etc.) by a significant margin.


\section{Conclusion}
\label{s:conclusion}
In this paper we have presented \emph{SwinDocSegmenter}, a powerful model to perform Document Layout Analysis by only utilizing the visual information. The improvement regarding the state-of-the-art is mainly constructed due to the enhanced and unified query selection, contrastive denoising training, and look forward twice approach. Also, the low-level projection head helps to enhance the low-level instances which makes a significant improvement in the overall performance as the Transformers are usually not good enough to detect small objects. However, there is still some scope for further improvement. The performance on PRIMA has still not reached the state-of-the-art with visual features as it contains a complex layout and very small training samples. A few-shot setting could help to improve the performance in the future. 

%
%
%
\section*{Acknowledgment}
This work has been partially supported by the Spanish project PID2021-126808OB-I00, the Catalan project 2021 SGR 01559 and the PhD Scholarship from AGAUR (2021FIB-10010). The Computer Vision Center is part of the CERCA Program / Generalitat de Catalunya.

\bibliographystyle{splncs04}
\bibliography{main}
\end{document}